\documentclass[conference]{IEEEtran}
\usepackage[mathlines,switch]{lineno}
\makeatletter
\let\old@ps@IEEEtitlepagestyle\ps@IEEEtitlepagestyle
\def\confheader#1{%
    \def\ps@IEEEtitlepagestyle{%
        \old@ps@IEEEtitlepagestyle%
        \def\@oddhead{\strut\hfill#1\hfill\strut}%
        \def\@evenhead{\strut\hfill#1\hfill\strut}%
    }%
    \ps@headings%
}
\makeatother
\confheader{%
        \parbox{20cm}{Accepted for publication in The International Joint Conference on Neural Networks (IJCNN), 2021}
}

\usepackage{centernot}
\usepackage{ragged2e}
\usepackage{lipsum}
\usepackage{microtype}
\usepackage{graphicx}
\usepackage{footmisc}
\usepackage{booktabs}
\usepackage{amsmath}
\usepackage{amsfonts}
\usepackage{amsbsy}
\usepackage{placeins}
\usepackage{siunitx}
\usepackage{bm}
\usepackage{comment}
\usepackage[utf8]{inputenc}
\usepackage[T1]{fontenc}
\usepackage{url}
\usepackage{nicefrac}
\usepackage{multirow}
\usepackage{mathtools}
\usepackage{xparse}
\DeclarePairedDelimiter{\norm}{\lVert}{\rVert}
\usepackage{empheq}

\usepackage{tabularx} 
\newcolumntype{C}{>{\centering\arraybackslash}X}
\ifCLASSOPTIONcompsoc
\usepackage[caption=false,font=normalsize,labelfont=sf,textfont=sf]{subfig}
\else
\usepackage[caption=false,font=footnotesize]{subfig}
\fi

\begin{document}

\title{Entropic Out-of-Distribution Detection}

\author{\IEEEauthorblockN{
David~Mac\^edo\,$^{1,2}$, 
Tsang~Ing~Ren\,$^{1}$,
Cleber~Zanchettin\,$^{1,3}$, %\\
Adriano~L.~I.~Oliveira\,$^{1}$, and
Teresa~Ludermir\,$^{1}$
}
\IEEEauthorblockA{\,
$^{1}$Centro de Inform\'atica, Universidade Federal de Pernambuco, Recife, Brasil\\
$^{2}$Montreal Institute for Learning Algorithms, University of Montreal, Quebec, Canada\\
$^{3}$Department of Chemical and Biological Engineering, Northwestern University, Evanston, United States of America\\
Emails: \{dlm, tir, cz, alio, tbl\}@cin.ufpe.br}
}

%\markboth{Journal of \LaTeX\ Class Files,~Vol.~14, No.~8, August~2015}{Shell \MakeLowercase{\textit{et al.}}: Bare Demo of IEEEtran.cls for IEEE Journals}
\markboth{Accepted for publication in The International Joint Conference on Neural Networks (IJCNN), 2021}{Mac\^edo~\MakeLowercase{\textit{et al.}}: Entropic Out-of-Distribution Detection}

\maketitle

\begin{abstract}
Out-of-distribution (OOD) detection approaches usually present special requirements (e.g., hyperparameter validation, collection of outlier data) and produce side effects (e.g., classification accuracy drop, slower energy-inefficient inferences). We argue that these issues are a consequence of the SoftMax loss anisotropy and disagreement with the maximum entropy principle. Thus, we propose the IsoMax loss and the entropic score. The seamless drop-in replacement of the SoftMax loss by IsoMax loss requires neither additional data collection nor hyperparameter validation. The trained models do not exhibit classification accuracy drop and produce fast energy-efficient inferences. Moreover, our experiments show that training neural networks with IsoMax loss significantly improves their OOD detection performance. The IsoMax loss exhibits state-of-the-art performance under the mentioned conditions (fast energy-efficient inference, no classification accuracy drop, no collection of outlier data, and no hyperparameter validation), which we call the seamless OOD detection task. In future work, current OOD detection methods may replace the SoftMax loss with the IsoMax loss to improve their performance on the commonly studied non-seamless OOD detection problem.
\end{abstract}

%\begin{IEEEkeywords}
%Document Segmentation, Octave Convolution, U-Net, Fully Octave Convolutional Network, FOCN, FOCNN.
%\end{IEEEkeywords}

\IEEEpeerreviewmaketitle

\begin{figure*}[!t]
\centering
\subfloat[]{\includegraphics[width=0.45\textwidth,trim={0 0 0 0},clip]{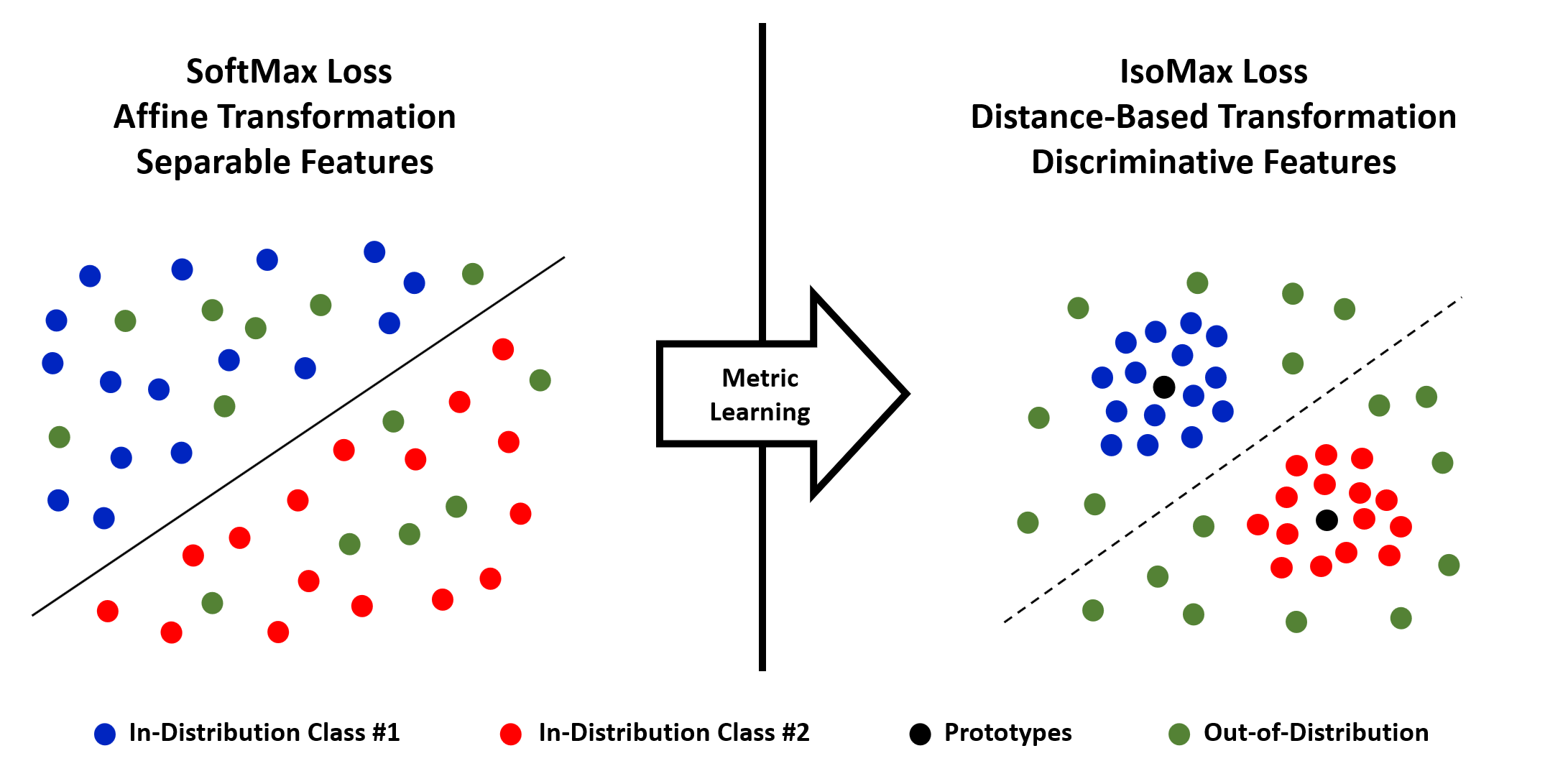}}\label{fig:separable-discriminative}
%\vskip -0.1cm
\subfloat[]{\includegraphics[width=0.53\textwidth,trim={0 0 0 0},clip]{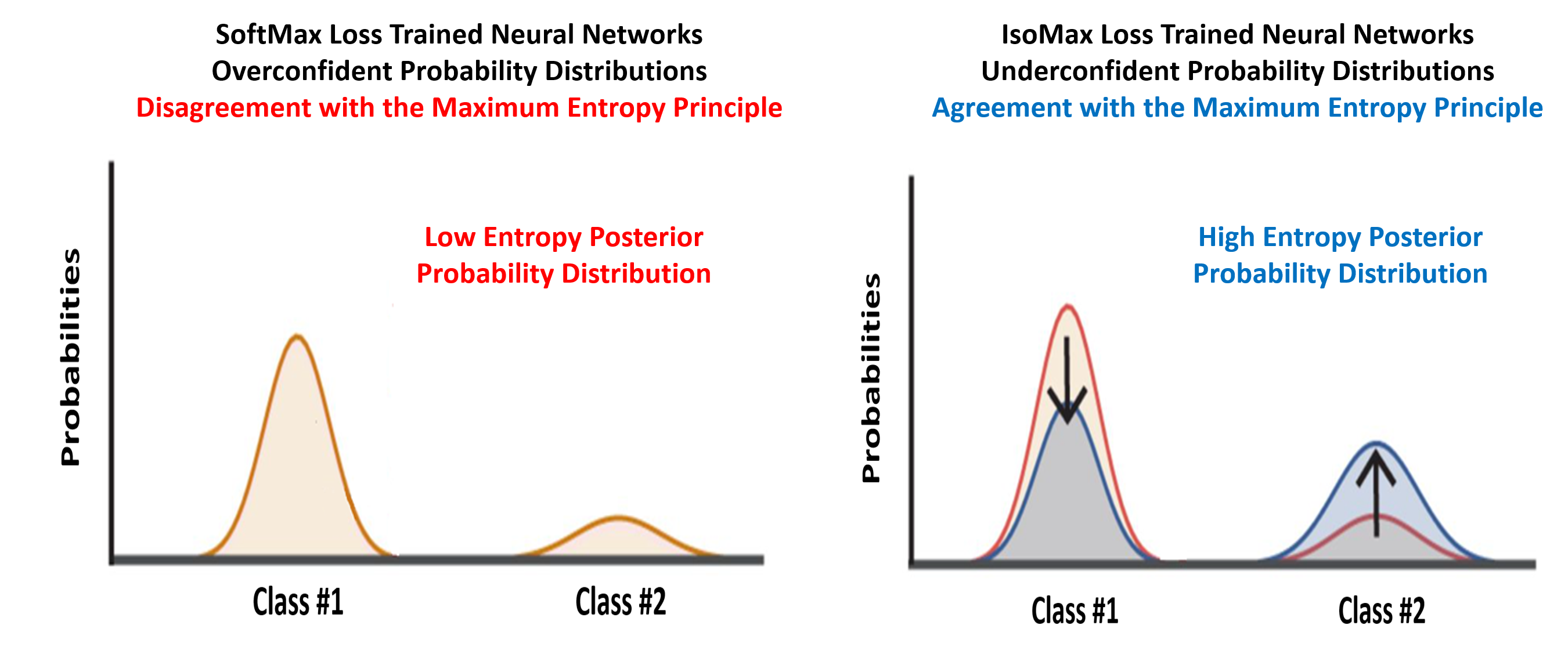}}\label{fig:from-softmax-to-isomax}
%\vskip -0.1cm
\caption{SoftMax loss drawbacks and IsoMax loss benefits: (a) Adapted from \cite[Fig.~1]{DBLP:conf/eccv/WenZL016}. SoftMax loss produces separable features \cite{DBLP:conf/eccv/WenZL016}. Postprocessing metric learning on features extracted from SoftMax loss-trained networks may convert from the situation on the left to the situation on the right \cite{lee2018simple, Mensink2013DistanceBasedIC, Scheirer_2013_TPAMI, Scheirer_2014_TPAMIb, 7298799,Rudd_2018_TPAMI}. The IsoMax loss, which is an exclusively distance-based (isotropic) loss, tends to generate more discriminative features \cite{DBLP:conf/eccv/WenZL016}. No feature extraction and subsequent metric learning are required when the IsoMax loss is used for training. (b) SoftMax loss trained neural networks produce overconfident low-entropy posterior probability distributions in disagreement with the maximum entropy principle. Our \emph{entropy maximization trick}, which consists in training using logits multiplied by a constant factor called the entropic scale \emph{that is nevertheless removed before inference}, enables IsoMax to generate underconfident high-entropy (almost maximum entropy) posterior probability distributions in agreement with the principle of maximum entropy.}
\label{fig:main}
\end{figure*}

\section{Introduction}

Out-of-distribution (OOD) detection approaches usually use \emph{special requirements} such as input preprocessing \cite{DeVries2018LearningNetworks, Hsu2020GeneralizedOD}, feature extraction combined with metric learning \cite{lee2018simple}, adversarial training \cite{Hein2018WhyRN}, hyperparameter validation \cite{liang2018enhancing}, and collection of additional data \cite{NIPS2018_8129, hendrycks2018deep, papadopoulos2019outlier, DBLP:journals/corr/abs-2010-03759}. Moreover, current OOD methods commonly show \emph{side effects} such as classification accuracy drop \cite{techapanurak2019hyperparameterfree, Hsu2020GeneralizedOD}, and slow energy-inefficient inferences \cite{liang2018enhancing, lee2018simple}. Solutions based on uncertainty (or confidence) estimation (or calibration) present complexity and lead to slow computationally inefficient inferences \cite{kendall2017uncertainties, Leibig2017LeveragingUI, subramanya2017confidence, malinin2018predictive, kuleshov2018accurate}.

We define the \emph{seamless} OOD detection task, which consists of performing OOD detection under the following restrictions. First, no classification accuracy drop is allowed. Second, the resulting models should produce inferences with the same speed and energy efficiency as those produced by the regularly trained neural networks. Third, no OOD/outlier/additional/extra data may be used. Finally, no hyperparameter validation is required. Improving the performance of neural networks in the \emph{seamless} OOD detection problem is important from a practical perspective. Additionally, such approaches may be combined in future work with current and novel OOD detection techniques to further improve the performance on the \emph{non-seamless} OOD detection task.

%To tackle the \emph{seamless} OOD detection problem, 
We argue that the unsatisfactory OOD detection performance of modern neural networks is mainly due to the drawbacks of the SoftMax loss (we follow the ``SoftMax loss'' expression as defined in \cite{liu2016large}). First, the SoftMax loss anisotropy does not incentivize the concentration of high-level representations in the feature space~\cite{DBLP:conf/eccv/WenZL016,Hein2018WhyRN}, making OOD detection difficult~\cite{Hein2018WhyRN} (Fig.~\ref{fig:main}a). Second, SoftMax loss produces overconfident low-entropy posterior probability distributions \cite{Guo2017OnCO}, which is in disagreement with the maximum entropy principle \cite{PhysRev.106.620, PhysRev.108.171, 10.5555/1146355} (Fig.~\ref{fig:main}b). Therefore, we propose the isotropy maximization loss \mbox{(IsoMax loss)}. To fix the SoftMax loss anisotropy, we made \mbox{IsoMax} an isotropic, i.e., exclusively distance-based, loss. To tackle the SoftMax loss overconfidence, we developed the \emph{entropy maximization trick}, which consists of training with logits multiplied by a high constant that is \emph{removed} for inference. This technique allows IsoMax loss to produce high-entropy (almost maximum) posterior probability distributions in agreement with the principle of maximum entropy.

%Rather than coexist with and circumvent SoftMax loss limitations, 
We propose to train neural networks replacing the SoftMax loss with the IsoMax loss. The swap of the SoftMax loss with the IsoMax loss requires changes in neither the architecture of the model nor training procedures or parameters. For OOD detection, we use the negative entropy of the neural network output probabilities, which we call the entropic score (ES). Since our solution presents neither \emph{special requirements} nor \emph{side effects}, it qualifies as a \emph{seamless} OOD detection approach as previously defined.

Our contributions are the following. First, we associate the unsatisfactory OOD detection performance of neural networks with the SoftMax loss anisotropy and disagreement with the maximum entropy principle. Second, we propose the IsoMax loss that acts as a \emph{SoftMax loss drop-in replacement} and may be used as a \emph{baseline for building improved OOD detection approaches} in future work. We show that the ES produces high performance combined with IsoMax loss. Third, we present the theoretical insight that associates the improved OOD detection performance of the networks trained with IsoMax loss with the principle of maximum entropy. Fourth, we show that our solution produces \emph{state-of-the-art} performance for the \emph{seamless} (fast energy-efficient inferences, no classification accuracy drop, no hyperparameter tuning, and no collection of outlier data) OOD detection task. Fifth, despite being unfair since the approaches present different \emph{special requirements} and \emph{side effects}, we compare our \emph{seamless} solution with \emph{non-seamless} OOD detection methods.

%\textbf{This paper is organized as follows: In Section II, we shortly present previous works tackled with similar problems. In Section III, we present our proposed methods. Section IV describes the experiment's setup, and Section V presents the results and discussion. In Section VI, presents our conclusions and possible future research.}

\section{Related Works}\label{sec:background}

Liang et al. \cite{liang2018enhancing} proposed the Out-of-DIstribution detector for Neural networks (ODIN) by combining input preprocessing and temperature calibration. The authors used OOD examples to validate the hyperparameters. Lee at al. \cite{lee2018simple} proposed the Mahalanobis distance-based approach by adding feature extraction, feature ensembles, and metric learning to input preprocessing. The authors proposed using adversarial examples rather than the OOD samples to tune hyperparameters. Since this procedure produces more realistic estimations, in this work, we only consider validation on adversarial samples for \emph{non-seamless} OOD detection methods. Hein at al. \cite{Hein2018WhyRN} proposed adversarial confidence enhancing training (ACET), which uses adversarial training. 

Hsu et al. \cite{Hsu2020GeneralizedOD} proposed to use the \emph{in-distribution} validation set for hyperparameter tuning. The CIFAR10/100 validation sets were used both for hyperparameter validation and for constructing the OOD detection test sets, which may have produced overestimated results. We believe that the in-distribution validation data used for defining hyperparameters should have been removed from the training set, which would presumably lead to an even stronger classification accuracy drop and, consequently, a decrease in OOD detection performance. The solution used input preprocessing and presented classification accuracy drop of a few percentage points in some cases. 

Techapanurak et al. \cite{techapanurak2019hyperparameterfree} used cosine similarity and a learnable block composed of batch normalization, an exponential, and a linear layer. The trained models presented classification accuracy drop of a few percentage points in some cases. Thus, the authors suggested using two models: one for classification and the other for OOD detection.

Recent approaches \cite{NIPS2018_8129,hendrycks2018deep,papadopoulos2019outlier,DBLP:journals/corr/abs-2010-03759} increased the OOD performance by training/fine-tuning using outlier data and hyperparameter validation. Liu et al. \cite{DBLP:journals/corr/abs-2010-03759} proposed the energy score. Methods based on uncertainty/confidence estimation/calibration \cite{kendall2017uncertainties,subramanya2017confidence,Leibig2017LeveragingUI,malinin2018predictive,kuleshov2018accurate} have been proposed to tackle the OOD detection problem.

\section{Entropic Out-of-Distribution Detection}\label{sec:seamless_approach}

\subsection{Isotropy.}

To fix the SoftMax loss anisotropy caused by its affine transformation, we forced the logits of the IsoMax loss to depend exclusively on the distances from the high-level features to the class prototypes.

Let $\bm{f}_{\bm{\theta}}(\bm{x})$ represent the high-level feature (embedding) associated with $\bm{x}$, $\bm{p}_{\bm{\phi}}^j$ represent the learnable prototype associated with class $j$, and $d()$ represent the \emph{nonsquared} distance. Additionally, let $\hat{y}^{(k)}$ represent the label of the correct class. Therefore, we construct an isotropic loss by writing:

%\begin{multline}
\begin{align}\label{eq:isotropic_loss2}
\begin{split}
\mathcal{L}_{I}(\hat{y}^{(k)}|\bm{x})
=-\log\left(\frac{\exp(-d(\bm{f}_{\bm{\theta}}(\bm{x}),\bm{p}_{\bm{\phi}}^k))}{\sum\limits_j\exp(-d(\bm{f}_{\bm{\theta}}(\bm{x}),\bm{p}_{\bm{\phi}}^j))}\right)
%\end{multline}
\end{split}
\end{align}

Unlike metric learning-based OOD detection approaches, rather than learning a metric from a preexisting feature space, in our solution, we learn a feature space that is from the start consistent with the chosen metric, avoiding the need for feature extraction and metric learning postprocessing phases after the neural network training.

\subsection{Entropy Maximization.}

Isotropy improves the OOD detection performance. However, for further performance gains, we need to circumvent the SoftMax loss propensity to produce low-entropy posterior probability distributions. To achieve high-entropy (almost maximum entropy) distributions in agreement with the maximum entropy principle, we introduce the entropic scale, which consists of a constant scalar factor applied to the logits \emph{throughout the training} that is nevertheless \emph{removed prior to inference}. We call this procedure the \emph{entropy maximization trick}. 

The entropic scale is equivalent to the \emph{inverse of the temperature} of the SoftMax \emph{function} (we follow the SoftMax \emph{function} expression as defined in \cite{liu2016large}). However, training with a \emph{predefined constant} entropic scale and then \emph{removing it before inference} is different from temperature calibration. On the one hand, the temperature of a \emph{pretrained} model is validated \emph{after training} and requires access to the OOD or adversarial examples. Furthermore, overoptimistic performance estimation is commonly produced~\cite{shafaei2018biased}. On the other hand, our approach requires neither hyperparameter validation nor access to the OOD or adversarial data. Rather than be applied to \emph{pretrained} models, our approach is used to train neural networks.

The presence of the entropic scale during training does not prevent the loss from approaching zero as required. However, when we remove it prior to the inference, the SoftMax \emph{function} naturally makes the entropy of the output probabilities increase to almost the maximum value possible if we use a high enough entropic score during training. Thus, returning to Equation \eqref{eq:isotropic_loss2}, multiplying the embedding-prototype distances by an entropic scale $E_s$, and representing the 2-norm of a vector by $\norm{.}$, we write the definition of the IsoMax loss as:

%\begin{multline}
\begin{align}\label{eq:loss_isomax}
\begin{split}
\mathcal{L}_{I}(\hat{y}^{(k)}|\bm{x})
=-\log\left(\frac{\exp(-E_s\norm{\bm{f}_{\bm{\theta}}(\bm{x})\!-\!\bm{p}_{\bm{\phi}}^k})}{\sum\limits_j\exp(-E_s\norm{\bm{f}_{\bm{\theta}}(\bm{x})\!-\!\bm{p}_{\bm{\phi}}^j})}\right)\\
%=-\log\left(\frac{\exp(-E_s\sqrt{(\bm{f}_{\bm{\theta}}(\bm{x})\!-\!\bm{p}_{\bm{\phi}}^k)\!\cdot\!(\bm{f}_{\bm{\theta}}(\bm{x})\!-\!\bm{p}_{\bm{\phi}}^k)})}{\sum\limits_j\exp(-E_s\sqrt{(\bm{f}_{\bm{\theta}}(\bm{x})\!-\!\bm{p}_{\bm{\phi}}^j)\!\cdot\!(\bm{f}_{\bm{\theta}}(\bm{x})\!-\!\bm{p}_{\bm{\phi}}^j)})}\right)
%\end{multline}
\end{split}
\end{align}

By applying the \emph{entropy maximization trick}, the inference probabilities for the IsoMax loss may be written as follows:

%\begin{multline}
\begin{align}\label{eq:probability_isomax}
\begin{split}
p_{I}(y^{(i)}|\bm{x})
&=\frac{\exp(-\norm{\bm{f}_{\bm{\theta}}(\bm{x})\!-\!\bm{p}_{\bm{\phi}}^i})}{\sum\limits_j\exp(-\norm{\bm{f}_{\bm{\theta}}(\bm{x})\!-\!\bm{p}_{\bm{\phi}}^j})}\\
%&=\frac{\exp(-\sqrt{(\bm{f}_{\bm{\theta}}(\bm{x})\!-\!\bm{p}_{\bm{\phi}}^k)\!\cdot\!(\bm{f}_{\bm{\theta}}(\bm{x})\!-\!\bm{p}_{\bm{\phi}}^k)})}{\sum\limits_j\exp(-\sqrt{(\bm{f}_{\bm{\theta}}(\bm{x})\!-\!\bm{p}_{\bm{\phi}}^j)\!\cdot\!(\bm{f}_{\bm{\theta}}(\bm{x})\!-\!\bm{p}_{\bm{\phi}}^j)})}
%\end{multline}
\end{split}
\end{align}

\subsection{Prototype Initialization.}

We observed that using the Xavier \cite{glorot2010understanding} or Kaiming \cite{He2016DelvingClassification} initializations for the prototypes leads to oscillations in performance. \emph{Hence, we decided to initialize all prototypes to the zero vector. Weight decay is applied to the prototypes because they are trainable parameters}.

\subsection{Entropic Score.}

The entropy has been studied for % anomaly detection \cite{DBLP:journals/csur/ChandolaBK09} and 
OOD detection \cite{DBLP:conf/nips/RenLFSPDDL19}. We show that the output probabilities negative entropy, which we call the entropic score, produces high-performance results when combined with IsoMax loss. 
\begin{comment}
%\begin{equation}\label{eq:entropy}
\begin{align}\label{eq:entropic_score}
\mathcal{ES}\!=\!-\sum_{i=1}^{N}{p(y^{(i)}|\bm{x})}\log p(y^{(i)}|\bm{x})
%\end{equation}
\end{align}
\end{comment}
Indeed, in such cases, the solution may consider the information provided by all network outputs rather than merely one output. For instance, ODIN and ACET only use the maximum probability.%, while the Mahalanobis method \cite{lee2018simple} only uses the distance to the nearest prototype.

\subsection{Implementation Details.}

To calculate the losses based on cross-entropy, deep learning libraries usually combine the logarithm and probability into a single computation. \emph{However, we experimentally observed that sequentially computing these calculations as standalone operations improves the IsoMax performance}. The class prototypes have the same dimension as the neural network last-layer representations. The number of prototypes is equal to the number of classes. The IsoMax loss has fewer parameters than the SoftMax loss because it has no bias to learn.

We observed classification accuracy drop and low/oscillating performance when trying to integrate the entropic scale into the SoftMax loss or with cosine similarity \cite{liu2016large, DBLP:journals/spl/WangCLL18, DBLP:conf/cvpr/DengGXZ19}. In such cases, the above strategy, i.e., initialization of the loss weights with the zero vector, cannot be performed. \emph{The Mahalanobis distance cannot be used because the covariance matrix is not differentiable}. Hence, the \emph{nonsquared Euclidean distance} is the optimal choice for integration with the entropic scale.

%\begin{comment}
\begin{table}[!t]
\small
\renewcommand{\arraystretch}{1.1}
\setlength{\tabcolsep}{0pt}
%\small
\centering
\caption[caption]{Classification Accuracy.}
%\vskip 0.1cm
\label{tab:classification_accuracy}
\begin{tabularx}{\columnwidth}{CC|CC}
\toprule
\multirow{2}{*}{Model} &
\multirow{2}{*}{Data}
& \multicolumn{2}{c}{Test Accuracy (\%) [$\uparrow$]}\\
&& SoftMax Loss & IsoMax Loss\\
\midrule
& SVHN & 96.6 & 96.6\\
DenseNet & CIFAR10 & 95.4 & 95.2\\
& CIFAR100 & 77.5 & 77.5\\
\midrule%{2-5}
& SVHN & 96.8 & 96.8\\
ResNet & CIFAR10 & 95.5 & 95.6\\
& CIFAR100 & 77.4 & 77.3\\
\bottomrule
\end{tabularx}
\vskip 0.01cm
\justify{In addition to avoiding classification accuracy drop compared with the SoftMax loss trained networks, IsoMax loss trained models show higher OOD detection performance (see Table~\ref{tbl:expanded_fair_odd}).}
\end{table}
%\end{comment}

\begin{figure*}[!t]
%\vskip -0.25 cm
\centering
\subfloat[]{\includegraphics[width=0.25\textwidth]{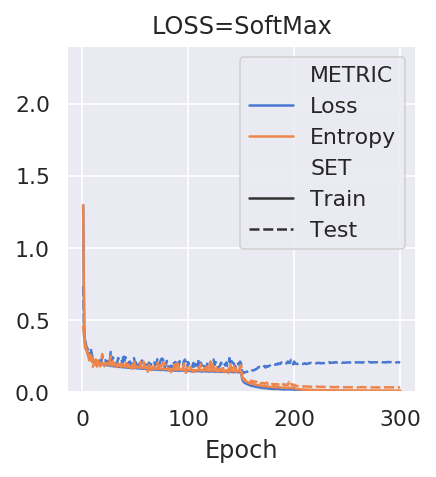}\label{fig:train_loss_entropies_softmax}}
\subfloat[]{\includegraphics[width=0.25\textwidth]{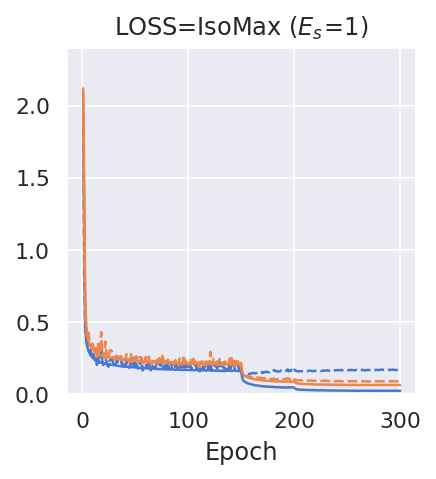}\label{fig:train_loss_entropies_isomax1}}
\subfloat[]{\includegraphics[width=0.25\textwidth]{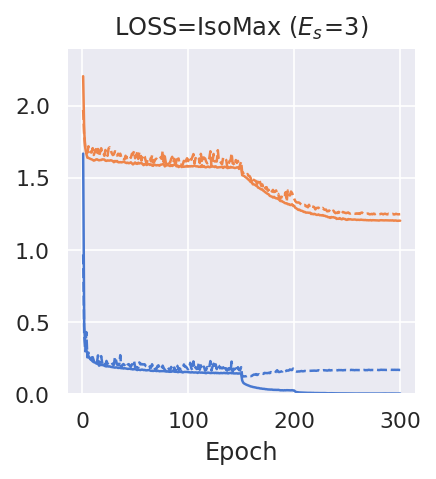}\label{fig:train_loss_entropies_isomax3}}
\subfloat[]{\includegraphics[width=0.25\textwidth]{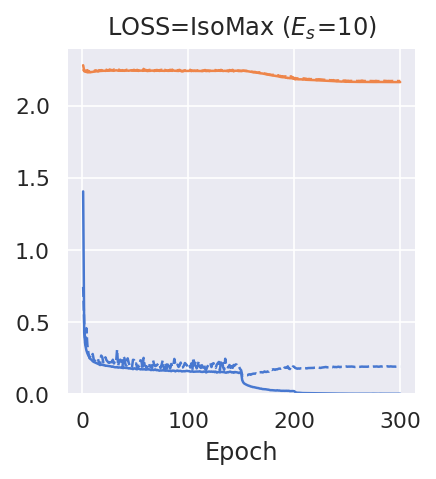}\label{fig:train_loss_entropies_isomax10}}
\vskip -0.05cm
%\\
%\vskip -0.05cm
\subfloat[]{\includegraphics[width=\textwidth]{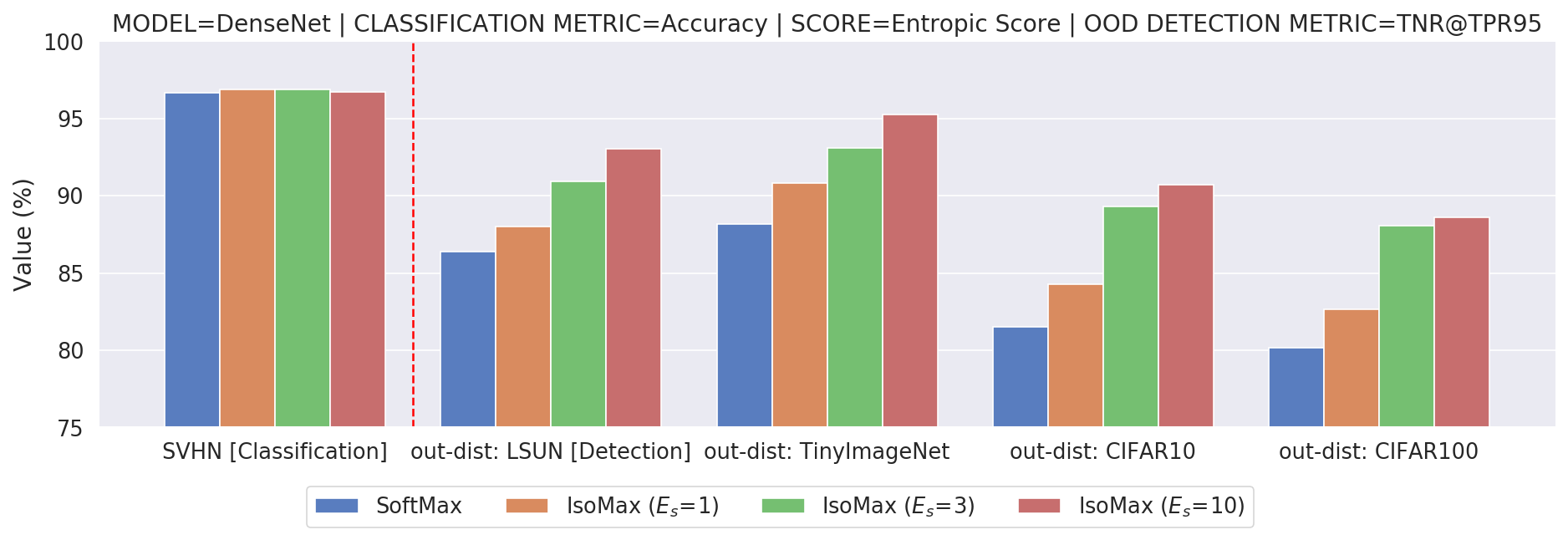}\label{fig:entropic_scale_parametrization}}
%\vskip -0.05cm
\caption{(a) SoftMax loss minimizes both the cross-entropy and the mean entropy of the posterior probabilities. (b)~IsoMax loss produces low mean entropy posterior probabilities for a low entropic scale \mbox{($E_s\!\!=\!1$)}. (c) IsoMax loss produces medium mean entropy for an intermediate entropic scale \mbox{($E_s\!\!=\!3$)}. (d) IsoMax loss produces high mean entropy for a high entropic scale \mbox{($E_s\!\!=\!10$)}. Therefore, higher entropic scale values are correlated with higher mean entropies as recommended by the maximum entropy principle. Notice that the orange line is almost flat in (d), so the IsoMax loss almost retains the maximum entropy  present at the beginning of the training for a high entropic scale. Hence, an entropic scale equal to ten is enough to produce posterior probability distributions with virtually the maximum possible mean entropy $\log(N)$, where $N$ is the number of classes. Consequently, there is no need to increase $E_s$ further. Therefore, we decided to use \mbox{$E_s\!\!=\!10$} for IsoMax loss (see also Fig.~\ref{fig:entropic_score_study}). (e) The left side of the dashed vertical red line presents the classification accuracies. The right side of the dashed vertical red line shows the OOD detection performance using the entropic score and the TNR@TPR95 (true negative rate at 95\% true positive rate) metric. We observe that a higher mean entropy produces increased OOD detection performance regardless of the out-of-distribution (out-dist). Isotropy by itself enables the IsoMax loss to exhibit higher performance than the SoftMax loss (\mbox{$E_s\!\!=\!1$}). IsoMax loss trained models exhibit classification accuracies similar to the classification accuracies presented by SoftMax loss trained networks regardless of the entropic scale.}
\label{fig:train_losses_entropies_and_entropic_scale_parametrization}
\end{figure*}

\begin{figure*}[!t]
\centering
\includegraphics[width=\textwidth]{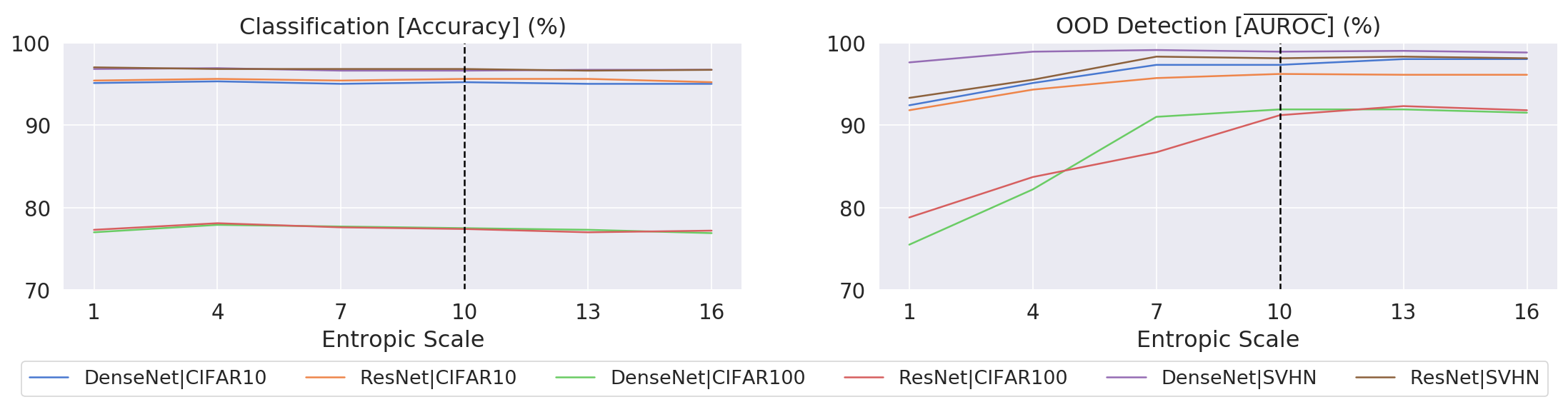}
%\vskip -0.25cm
\caption{$\overline{\mathrm{AUROC}}$ represents the mean AUROC considering all out-of-distribution data. The classification accuracy and the mean OOD detection performance are approximately stable for \mbox{$E_s\!\!=\!10$} or higher regardless of the dataset and model. $E_s$ validation cannot significantly improve the OOD detection performance. In fact, this is not even possible because access to the OOD or outlier samples is not allowed in \emph{seamless} OOD detection. Making $E_s$ learnable did not considerably improve or decrease the OOD detection results.}
\label{fig:entropic_score_study}
\end{figure*}

\begin{table*}[!t]
\small
\renewcommand{\arraystretch}{0.5}
\setlength{\tabcolsep}{0pt}
%\small
\centering
\caption[caption]{Seamless Out-of-Distribution Detection: %Fair comparison of approaches with the same requirements and side effects.
No hyperparameter tuning. Fast and energy-efficient inferences.\\No classification accuracy drop. No outlier/background data.}
%\vspace{-0.1in}
\begin{tabularx}{\textwidth}{lll|CCC}
\toprule
\multirow{4}{*}{\begin{tabular}[c]{@{}c@{}}\\Model\end{tabular}} & \multirow{4}{*}{\begin{tabular}[c]{@{}c@{}}\\Data\\(training)\end{tabular}} & \multirow{4}{*}{\begin{tabular}[c]{@{}c@{}}\\OOD\\(unseen)\end{tabular}} & 
\multicolumn{3}{c}{Seamless OOD Detection: No Classification Accuracy Drop. No Outlier Data.}\\
&&& \multicolumn{3}{c}{Fast and Energy-Efficient Inferences. No Hyperparameter Tuning.}\\
\cmidrule{4-6}
&&& TNR@TPR95\textsuperscript{1} (\%) [$\uparrow$] & AUROC\textsuperscript{2} (\%) [$\uparrow$] & DTACC\textsuperscript{3} (\%) [$\uparrow$]\\
&&& \multicolumn{3}{c}{SoftMax+MPS\textsuperscript{4} / SoftMax+ES\textsuperscript{5} / IsoMax+MPS\textsuperscript{6} / IsoMax+ES\textsuperscript{7} (ours)}\\
\midrule
\multirow{9}{*}{\begin{tabular}[c]{@{}c@{}}DenseNet~~~~~~\end{tabular}}
& \multirow{3}{*}{\begin{tabular}[c]{@{}c@{}}CIFAR10~~~~~~~\end{tabular}} 
& SVHN & 32.2 / 33.2 / 64.5 / \bf77.0 & 86.6 / 86.9 / 94.6 / \bf96.6 & 79.9 / 79.9 / 88.1 / \bf91.6\\
&& TinyImageNet~\cite{Deng2009ImageNetDatabase}~~~~~~~~& 55.8 / 59.8 / 81.1 / \bf88.0 & 93.5 / 94.2 / 96.8 / \bf97.8 & 87.6 / 87.8 / 90.8 / \bf93.2\\
&& LSUN~\cite{Yu2015LSUNLoop} & 64.9 / 69.5 / 88.5 / \bf94.5 & 95.2 / 95.9 / 97.9 / \bf98.8 & 89.9 / 90.0 / 93.1 / \bf94.9\\
\cmidrule{2-6} 
& \multirow{3}{*}{\begin{tabular}[c]{@{}c@{}}CIFAR100\end{tabular}} 
& SVHN & 20.6 / 24.9 / {\bf27.5} / 23.4 & 80.1 / 81.9 / 86.3 / \bf88.6 & 73.9 / 74.3 / 79.9 / \bf83.7\\
&& TinyImageNet & 19.4 / 23.7 / 42.4 / \bf49.1 & 77.0 / 78.8 / 90.2 / \bf92.6 & 70.6 / 71.1 / 83.6 / \bf86.6\\
&& LSUN & 18.8 / 24.4 / 48.9 / \bf63.0 & 75.9 / 77.9 / 91.3 / \bf94.7 & 69.5 / 70.2 / 84.2 / \bf89.1\\
\cmidrule{2-6} 
& \multirow{3}{*}{\begin{tabular}[c]{@{}c@{}}SVHN\end{tabular}} 
& CIFAR10 & 81.5 / 83.7 / 91.6 / \bf94.1 & 96.5 / 96.9 / 98.2 / \bf98.5 & 91.9 / 92.1 / 94.1 / \bf95.0\\
&& TinyImageNet & 88.2 / 90.0 / 95.3 / \bf97.0 & 97.7 / 98.1 / 98.9 / \bf99.1 & 93.5 / 93.7 / 95.4 / \bf96.1\\
&& LSUN & 86.4 / 88.4 / 94.7 / \bf96.8 & 97.3 / 97.8 / 98.7 / \bf99.1 & 92.8 / 93.0 / 95.0 / \bf95.9\\
\midrule
\multirow{9}{*}{\begin{tabular}[c]{@{}c@{}}ResNet\end{tabular}}
& \multirow{3}{*}{\begin{tabular}[c]{@{}c@{}}CIFAR10\end{tabular}} 
& SVHN & 43.1 / 44.5 / 81.7 / \bf83.6 & 91.7 / 92.0 / 96.8 / \bf97.1 & 86.5 / 86.5 / 91.2 / \bf91.9\\
&& TinyImageNet & 46.3 / 48.0 / 66.0 / \bf70.2 & 89.8 / 90.0 / 93.9 / \bf94.6 & 84.0 / 84.1 / 87.1 / \bf88.3\\
&& LSUN & 51.2 / 53.3 / 76.6 / \bf82.3 & 92.2 / 92.6 / 96.2 / \bf96.9 & 86.5 / 86.6 / 90.1 / \bf91.5\\
\cmidrule{2-6} 
& \multirow{3}{*}{\begin{tabular}[c]{@{}c@{}}CIFAR100\end{tabular}} 
& SVHN & 15.9 / 18.0 / {\bf22.5} / 20.2 & 71.3 / 72.7 / 83.9 / \bf85.3 & 66.1 / 66.3 / 77.8 / \bf79.7\\
&& TinyImageNet & 18.5 / 22.4 / 38.9 / \bf50.6 & 74.7 / 76.3 / 89.2 / \bf92.0 & 68.8 / 69.1 / 82.2 / \bf85.6\\
&& LSUN & 18.4 / 22.4 / 41.4 / \bf54.8 & 74.7 / 76.5 / 90.1 / \bf93.2 & 69.1 / 69.4 / 83.3 / \bf87.5\\
\cmidrule{2-6} 
& \multirow{3}{*}{\begin{tabular}[c]{@{}c@{}}SVHN\end{tabular}} 
& CIFAR10 & 67.3 / 67.7 / 90.5 / \bf92.3 & 89.8 / 89.7 / 97.9 / \bf98.0 & 87.0 / 86.9 / 93.7 / \bf94.1\\
&& TinyImageNet & 66.9 / 67.3 / 92.1 / \bf94.4 & 89.0 / 89.0 / 98.2 / \bf98.4 & 86.7 / 86.6 / 94.3 / \bf94.8\\
&& LSUN & 62.2 / 62.5 / 88.6 / \bf90.8 & 86.0 / 85.8 / 97.6 / \bf97.8 & 84.2 / 84.1 / 93.4 / \bf93.6\\
\bottomrule
\end{tabularx}
\vskip 0.01cm
\justify{\textsuperscript{1}True negative rate at 95\% true positive rate. \textsuperscript{2}Area under the receiver operating
characteristic curve. \textsuperscript{3}Detection accuracy \cite{lee2018simple}. \textsuperscript{4}SoftMax+MPS means training with SoftMax loss and performing OOD detection using the maximum probability score (MPS), which is the approach defined in \cite{hendrycks2017baseline}. \textsuperscript{5}SoftMax+ES means training with SoftMax loss and performing OOD detection using the entropic score (ES). \textsuperscript{6}IsoMax+MPS means training with IsoMax loss and performing OOD detection using the maximum probability score (MPS). \textsuperscript{7}IsoMax+ES means training with IsoMax loss and performing OOD detection using the entropic score (ES) (our proposal). The best results are shown in bold. To the best of our knowledge, our IsoMax+ES approach presents \emph{state-of-the-art} performance under these severely restrictive assumptions.}
\label{tbl:expanded_fair_odd}
\end{table*}

\section{Experiments}\label{sec:experiments}

All datasets, models, and evaluation metrics used the baseline established in \cite{hendrycks2017baseline} that was followed in the major OOD detection papers \cite{liang2018enhancing,lee2018simple,Hein2018WhyRN}. We trained from the scratch 100-layer DenseNet-BC \cite{Huang2017DenselyNetworks} (growth rate $k\!=\!12$, 0.8M parameters) and 34-layer ResNets \cite{He_2016} on CIFAR10 \cite{Krizhevsky2009LearningImages}, CIFAR100 \cite{Krizhevsky2009LearningImages} and SVHN \cite{Netzer2011ReadingLearning} using the SoftMax and IsoMax losses.

For both the SoftMax loss and IsoMax loss, we used SGD with the Nesterov moment equal to 0.9, 300 epochs with a batch size of 64, and an initial learning rate of 0.1, with a learning rate decay rate equal to ten applied in epochs 150, 200, and 250. We used a dropout of zero. The weight decay was 0.0001.

We only compared approaches that did not present classification accuracy drop because this facilitates increasing OOD detection performance \cite{DBLP:conf/iclr/TsiprasSETM19}; moreover, it is particularly undesired from a practical perspective \cite{carlini2019evaluating}. It is well known that using OOD/outlier/background/additional data improves the OOD detection performance. Therefore, considering that data-based regularization techniques may benefit both SoftMax loss and IsoMax loss, we perform all experiments without outlier exposure \cite{hendrycks2018deep, papadopoulos2019outlier}, background samples \cite{NIPS2018_8129}, or energy-based fine-tuning \cite{DBLP:journals/corr/abs-2010-03759}. The source code is available online\footnote{\url{https://github.com/dlmacedo/entropic-out-of-distribution-detection}}.
%online\footnote{\url{https://codeocean.com/capsule/7175931/tree/v1}}.

\section{Results and Discussions}

\subsection{IsoMax Loss Properties, Ablation Study, and Entropic Scale Value Definition.}

To experimentally show that higher entropic scales lead to higher mean entropy probability distributions and consequently improve the OOD detection performance, we trained DenseNets on SVHN using the SoftMax loss and IsoMax loss with distinct entropic scale values. We used the entropic score and the TNR@TPR95 (true negative rate at 95\% true positive rate) to evaluate the OOD detection performance (Fig.~\ref{fig:train_losses_entropies_and_entropic_scale_parametrization}).

Fig.~\ref{fig:train_losses_entropies_and_entropic_scale_parametrization}a shows that the SoftMax loss generates posterior distributions with low mean entropy. Fig.~\ref{fig:train_losses_entropies_and_entropic_scale_parametrization}b illustrates that the unitary entropic scale \mbox{($E_s\!\!=\!1$)} does not increase the mean entropy of the probability distributions. In other words, isotropy alone is not enough to produce low mean entropy probability distributions, and the \emph{entropy maximization trick} is necessary. Nevertheless, Fig.~\ref{fig:train_losses_entropies_and_entropic_scale_parametrization}e shows that the simple replacement of anisotropic logits based on the affine transformation by isotropic logits is enough to produce some OOD detection performance gains for all out-of-distribution (out-dist) data, even without the mentioned trick \mbox{($E_s\!\!=\!1$)}. 
Fig.~\ref{fig:train_losses_entropies_and_entropic_scale_parametrization}c shows that an intermediate entropic scale \mbox{($E_s\!\!=\!3$)} provides medium mean entropy probability distributions with the additional OOD detection performance gains regardless of the out-of-distribution data (Fig.~\ref{fig:train_losses_entropies_and_entropic_scale_parametrization}e). Fig.~\ref{fig:train_losses_entropies_and_entropic_scale_parametrization}d illustrates that a high entropic scale \mbox{($E_s\!\!=\!10$)} produces even higher mean entropy probability distributions and the highest OOD detection performance for all out-of-distribution data considered (Fig.~\ref{fig:train_losses_entropies_and_entropic_scale_parametrization}e). We emphasize that regardless of training with the entropic scale, \emph{if it is not removed for inference}, the IsoMax loss produces outputs with entropies as low as those produced by the SoftMax loss and high OOD detection performances are no long observed. 

\begin{table*}[!t]
\small
\renewcommand{\arraystretch}{0.5}
\setlength{\tabcolsep}{0pt}
\centering
\caption[caption]{Non-seamless Out-of-Distribution Detection: Unfair comparison of approaches with different special requirements\\and side effects. No classification accuracy drop. No outlier/background data.}
\label{tbl:unfair_odd}
%\small
%\centering
%\vspace{-0.12in}
\begin{tabularx}{\textwidth}{lll|CC}
\toprule
\multirow{4}{*}{\begin{tabular}[c]{@{}c@{}}\\Model\end{tabular}} & \multirow{4}{*}{\begin{tabular}[c]{@{}c@{}}\\Data\\(training)\end{tabular}} & \multirow{4}{*}{\begin{tabular}[c]{@{}c@{}}\\OOD\\(unseen)\end{tabular}} & 
\multicolumn{2}{c}{Non-seamless Out-of-Distribution Detection:}\\
&&& \multicolumn{2}{c}{Approaches with Different Special Requirements and Side Effects.}\\
\cmidrule{4-5}
&&& AUROC (\%) [$\uparrow$] & DTACC\textsuperscript{5} (\%) [$\uparrow$]\\
&&& \multicolumn{2}{c}{ODIN\textsuperscript{1} / ACET\textsuperscript{2} / IsoMax+ES\textsuperscript{3} (ours) / Mahalanobis\textsuperscript{4}}\\
\midrule
\multirow{9}{*}{\begin{tabular}[c]{@{}c@{}}DenseNet~~~~~~~~~~~~\end{tabular}}
& \multirow{3}{*}{\begin{tabular}[c]{@{}c@{}}CIFAR10~~~~~~~~~~~~~\end{tabular}} 
& SVHN & 92.8 / NA / {\bf96.6} / \bf97.6 & 86.5 / NA / {\bf91.6} / \bf92.6\\
&& TinyImageNet~~~~~~~~~~~~& 97.2 / NA / {\bf97.8} / \bf98.8 & 92.1 / NA / {\bf93.2} / \bf95.0\\
&& LSUN & 98.5 / NA / {\bf98.8} / \bf99.2 & 94.3 / NA / {\bf94.9} / \bf96.2\\
\cmidrule{2-5} 
& \multirow{3}{*}{\begin{tabular}[c]{@{}c@{}}CIFAR100\end{tabular}} 
& SVHN & 88.2 / NA / 88.6 / \bf91.8 & 80.7 / NA / {\bf83.7} / \bf84.6\\
&& TinyImageNet & 85.3 / NA / 92.6 / \bf97.0 & 77.2 / NA / 86.6 / \bf91.8\\
&& LSUN & 85.7 / NA / 94.7 / \bf97.9 & 77.3 / NA / 89.1 / \bf93.8\\
\cmidrule{2-5} 
& \multirow{3}{*}{\begin{tabular}[c]{@{}c@{}}SVHN\end{tabular}} 
& CIFAR10 & 91.9 / NA / {\bf98.5} / \bf98.8 & 86.6 / NA / {\bf95.0} / \bf96.3\\
&& TinyImageNet & 94.8 / NA / {\bf99.1} / \bf99.8 & 90.2 / NA / 96.1 / \bf98.9\\
&& LSUN & 94.1 / NA / {\bf99.1} / \bf99.9 & 89.1 / NA / 95.9 / \bf99.2\\
\midrule
\multirow{9}{*}{\begin{tabular}[c]{@{}c@{}}ResNet\end{tabular}}
& \multirow{3}{*}{\begin{tabular}[c]{@{}c@{}}CIFAR10\end{tabular}} 
& SVHN & 86.5 / {\bf98.1} / {\bf97.1} / 95.5 & 77.8 / NA / {\bf91.9} / \bf89.1\\
&& TinyImageNet & 93.9 / 85.9 / 94.6 / \bf99.0 & 86.0 / NA / 88.3 / \bf95.4\\
&& LSUN & 93.7 / 85.8 / 96.9 / \bf99.5 & 85.8 / NA / 91.5 / \bf97.2\\
\cmidrule{2-5} 
& \multirow{3}{*}{\begin{tabular}[c]{@{}c@{}}CIFAR100\end{tabular}} 
& SVHN & 72.0 / {\bf91.2} / 85.3 / 84.4 & 67.7 / NA / {\bf79.7} / 76.5\\
&& TinyImageNet & 83.6 / 75.2 / {\bf92.0} / 87.9 & 75.9 / NA / {\bf85.6} / \bf84.6\\
&& LSUN & 81.9 / 69.8 / {\bf93.2} / 82.3 & 74.6 / NA / {\bf87.5} / 79.7\\
\cmidrule{2-5} 
& \multirow{3}{*}{\begin{tabular}[c]{@{}c@{}}SVHN\end{tabular}} 
& CIFAR10 & 92.1 / 97.3 / {\bf98.0} / {\bf97.6} & 89.4 / NA / {\bf94.1} / \bf94.6\\
&& TinyImageNet & 92.9 / 97.7 / {\bf98.4} / \bf99.3 & 90.1 / NA / 94.8 / \bf98.8\\
&& LSUN & 90.7 / {\bf99.7} / 97.8 / \bf99.9 & 88.2 / NA / 93.6 / \bf99.5\\
\bottomrule
\end{tabularx}
\vskip 0.01cm
\justify{ODIN, the Mahalanobis approach, and ACET present hyperparameters that must be validated for each combination of datasets and models. They also require previously known optimal adversarial perturbation values for each combination of datasets and models. \textsuperscript{1}ODIN uses input preprocessing, temperature calibration, and adversarial validation, i.e., hyperparameter tuning using adversarial examples \cite{liang2018enhancing}. \textsuperscript{2}ACET uses adversarial training, resulting in slower training, possibly reduced scalability for large images, and eventually classification accuracy drop \cite{Hein2018WhyRN}. \textsuperscript{3}IsoMax+ES means training with IsoMax loss and performing OOD detection using the entropic score (ES). Considering that validating $E_s$ using adversarial examples cannot produce significant gains (Fig.~\ref{fig:entropic_score_study}), we prefer to keep $E_s\!\!=\!10$ to maintain the simplicity of the solution. %We did \emph{not} used adversarial samples to validated $E_s$. Hence, we kept $E_s\!\!=\!10$ for all experiments.
\textsuperscript{4}The Mahalanobis solution uses input preprocessing, feature ensemble, feature extraction followed by metric learning, and adversarial validation~\cite{lee2018simple}. \textsuperscript{5}Detection accuracy \cite{lee2018simple}. The best results are shown in bold (2\% tolerance).}
\end{table*}

\begin{table*}[!t]
%\vskip -0.5cm
\small
\renewcommand{\arraystretch}{0.5}
\setlength{\tabcolsep}{4pt}
\centering
\caption{Non-seamless OOD Detection: Inference delays. Presumed computational cost and energy consumption rates.}
\label{tbl:times}
\begin{tabularx}{\textwidth}{lll|CCC}
\toprule
\multirow{4}{*}{\begin{tabular}[c]{@{}c@{}}\\Model\end{tabular}} & \multirow{4}{*}{\begin{tabular}[c]{@{}c@{}}\\Data\\(training)\end{tabular}} & \multirow{4}{*}{\begin{tabular}[c]{@{}c@{}}\\Hardware\\(inference)\end{tabular}} & 
\multicolumn{3}{c}{Non-seamless Out-of-Distribution Detection:}\\
&&& \multicolumn{3}{c}{Inference Delays. Presumed Computational Cost and Energy Consumption Rates.}\\
\cmidrule{4-6}
&&& SoftMax Loss~\cite{hendrycks2017baseline} & IsoMax Loss (ours) & \mbox{ODIN~\cite{liang2018enhancing}, Mahalanobis~\cite{lee2018simple}}, Generalized ODIN~\cite{Hsu2020GeneralizedOD}\\
&&& MPS / ES (ms) [$\downarrow$] & MPS / ES (ms) [$\downarrow$] & (ms) [$\downarrow$]\\
\midrule
\multirow{6}{*}{\begin{tabular}[c]{@{}c@{}}DenseNet~~~~~~\end{tabular}}
& \multirow{2}{*}{\begin{tabular}[c]{@{}c@{}}CIFAR10~~~~~~~\end{tabular}} 
&~~~~CPU~~~~~~& 18.1 / 19.4 & 18.0 / 19.2 & 242.4 \bf{($\approx$ 10x slower)}\\
&&~~~~GPU~~~~~~& 11.6 / 13.0 & 11.6 / 11.5 & 39.2 \bf{($\approx$ 4x slower)}\\
\cmidrule{2-6} 
& \multirow{2}{*}{\begin{tabular}[c]{@{}c@{}}CIFAR100\end{tabular}} 
&~~~~CPU~~~~~~& 18.4 / 19.8 & 18.4 / 19.3 & 261.0 \bf{($\approx$ 10x slower)}\\
&&~~~~GPU~~~~~~& 12.9 / 11.4 & 11.8 / 11.5 & 39.6 \bf{($\approx$ 4x slower)}\\
\cmidrule{2-6} 
& \multirow{2}{*}{\begin{tabular}[c]{@{}c@{}}SVHN\end{tabular}} 
&~~~~CPU~~~~~~& 18.1 / 18.6 & 18.3 / 18.6 & 241.5 \bf{($\approx$ 10x slower)}\\
&&~~~~GPU~~~~~~& 11.6 / 11.9 & 11.7 / 11.6 & 39.6 \bf{($\approx$ 4x slower)}\\
\midrule
\multirow{6}{*}{\begin{tabular}[c]{@{}c@{}}ResNet\end{tabular}}
& \multirow{2}{*}{\begin{tabular}[c]{@{}c@{}}CIFAR10\end{tabular}} 
&~~~~CPU~~~~~~& 22.3 / 23.2 & 23.0 / 23.5 & 250.4 \bf{($\approx$ 10x slower)}\\
&&~~~~GPU~~~~~~& 4.5 / 3.8 & 4.2 / 4.1 & 15.4 \bf{($\approx$ 4x slower)}\\
\cmidrule{2-6} 
& \multirow{2}{*}{\begin{tabular}[c]{@{}c@{}}CIFAR100\end{tabular}} 
&~~~~CPU~~~~~~& 23.3 / 23.1 & 23.3 / 23.8 & 252.6 \bf{($\approx$ 10x slower)}\\
&&~~~~GPU~~~~~~& 4.3 / 3.9 & 4.3 / 4.2 & 14.8 \bf{($\approx$ 4x slower)}\\
\cmidrule{2-6} 
& \multirow{2}{*}{\begin{tabular}[c]{@{}c@{}}SVHN\end{tabular}} 
&~~~~CPU~~~~~~& 23.1 / 23.4 & 23.4 / 23.3 & 263.8 \bf{($\approx$ 10x slower)}\\
&&~~~~GPU~~~~~~& 4.2 / 4.0 & 4.0 / 4.0 & 15.7 \bf{($\approx$ 4x slower)}\\
\bottomrule
\end{tabularx}
\vskip 0.01cm
\justify{MPS means maximum probability score. ES means entropic score. For SoftMax loss and IsoMax loss, the inference delays combine both classification and detection computation. For the methods based on input preprocessing, the inference delays represent only the input preprocessing phase. All values are in milliseconds. The inference delay rates presumably reflect similar computational cost and energy consumption rates.}
\end{table*}

Hence, the \emph{entropy maximization trick} enables the migration from low-entropy distributions (Fig.~\ref{fig:train_losses_entropies_and_entropic_scale_parametrization}a,b) to high-entropy distributions (Fig.~\ref{fig:train_losses_entropies_and_entropic_scale_parametrization}d). For a high entropic scale, the IsoMax loss minimizes the cross-entropy while producing high-entropy probability distributions as recommended by the principle of maximum entropy. More importantly, higher entropy posterior probability distributions directly correlate with increased OOD detection performances despite the out-of-distribution data. Fig.~\ref{fig:train_losses_entropies_and_entropic_scale_parametrization}d shows that an entropic scale \mbox{$E_s\!\!=\!10$} is enough to produce essentially the maximum possible entropy. Therefore, we defined the entropic scale as a \emph{constant} equal to ten.

\subsection{Classification Accuracy and OOD Detection Performance Dependence on the \mbox{Entropic} Scale Value.}

\emph{After defining \mbox{$E_s\!\!=\!10$} for the IsoMax loss based on the previous experiments}, we performed additional analyses. Fig.~\ref{fig:entropic_score_study} shows that \emph{regardless of the combination of dataset and model}, the classification accuracy and the mean OOD detection performance are essentially stable for \mbox{$E_s\!\!=\!10$} or higher, as the entropic scale is already high enough to ensure near-maximal entropy. Hence, validation of $E_s$ does not produce a considerable performance increase. In fact, this is not even possible because we consider access to OOD or outlier samples to be forbidden. Making $E_s$ learnable did not significantly affect the OOD detection performance. Table~\ref{tab:classification_accuracy} shows that the IsoMax loss trained models do not show classification accuracy drop.

\subsection{Seamless Out-of-Distribution Detection.}

To the best of our knowledge, the proposal presented in \cite{hendrycks2017baseline} and our method are the only solutions that qualify as \emph{seamless} OOD detection approaches. Table~\ref{tbl:expanded_fair_odd} shows that the models trained with the SoftMax loss using the maximum probability as the score \mbox{(SoftMax+MPS)} always present the worst performance~results and that replacing the maximum probability score by the entropic score (SoftMax+ES) produces OOD detection performance gains.

The combination of the models trained using IsoMax loss with the entropic score (IsoMax+ES), which is the proposed solution, significantly improves, usually by several percentage points, the OOD detection performance across almost all datasets, models, out-of-distribution data, and metrics.

%The entropic score was used without much success \cite{DBLP:conf/nips/RenLFSPDDL19}, our experiments show that 
The entropic score produces high OOD detection performance when the distributions present high entropy (IsoMax+ES). Indeed, both \emph{producing high-entropy distributions} and the \emph{entropic score} contribute to improving the OOD detection performance. However, the contribution of \emph{producing high-entropy distributions is considerably more important}. 

The model does not affect the analyses presented. Indeed, the comments shown above are valid for both DenseNet and ResNet models.

\subsection{Non-seamless Out-of-Distribution Detection.}

To tackle \emph{non-seamless} OOD detection, IsoMax should work as a \emph{baseline} to be combined with OOD techniques (e.g., outlier exposure, adversarial training, input preprocessing, energy score) rather than competing as a standalone solution. Nevertheless, Table~\ref{tbl:unfair_odd} provides a perspective for how our \emph{baseline seamless (standalone)} approach compares to \emph{non-seamless (composed)} solutions.

From a qualitative perspective, ODIN and Mahalanobis use input preprocessing; i.e., to perform OOD detection, each inference requires a first neural network forward pass, a backpropagation, and a second forward pass. They produce slower and less energy-efficient inferences than models trained with IsoMax loss, which are as fast and computationally efficient as the models trained with SoftMax loss. Input preprocessing is indeed a limitation from an economic and environmental perspective \cite{Schwartz2019GreenA}.

ODIN requires temperature calibration after neural network training, while the Mahalanobis approach requires feature ensemble and metric learning. Unlike pretraining-based solutions, our approach requires no postprocessing after the neural network training. ACET requires adversarial training, which produces slower training and may limit the application of ACET to large images \cite{DBLP:conf/nips/ShafahiNG0DSDTG19}.

From a quantitative point of view, Table~\ref{tbl:unfair_odd} shows that IsoMax+ES considerably outperforms ODIN in all evaluated scenarios. Therefore, in addition to avoiding hyperparameter tuning and access to the OOD or adversarial samples, the results show that the \emph{entropy maximization trick} is much more effective in improving the OOD detection performance than temperature calibration, even when the latter is combined with input preprocessing. Furthermore, IsoMax+ES usually outperforms ACET, in some cases by a large margin. Moreover, in most cases, the Mahalanobis method surpasses IsoMax+ES by less than 2\%. In some scenarios, IsoMax+ES outperforms the Mahalanobis method.

Table~\ref{tbl:times} presents the inference delays for the SoftMax loss, IsoMax loss, and competing methods using a CPU and GPU. We observe that neural networks trained using the IsoMax loss produce inferences equally as fast as those produced by networks trained using the SoftMax loss, regardless of whether a CPU or GPU is used for inference.

Additionally, the entropic score is as fast as the usual maximum probability score. Moreover, the methods based on input preprocessing were more than ten times slower on the CPU and approximately four times slower on the GPU. These ratios also presumably apply to the computational cost and energy consumption.

To agree with the maximum entropy principle and achieve high performance, rather than generating \emph{calibrated} maximum probabilities, IsoMax must produce the \emph{lowest possible} maximum probabilities.

\section{Conclusion}\label{sec:conclusion}

We proposed a \emph{seamless} OOD detection approach based on logit isotropy and the maximum entropy principle. The proposed IsoMax loss acts as a SoftMax loss drop-in replacement that produces accurate predictions in addition to fast energy- and computation-efficient inferences. No hyperparameter tuning is needed. Hence, no additional procedure other than straightforward neural network training is needed.

OOD detection is performed using the rapid entropic score. Collection of outlier/background data is also not required. To the best of our knowledge, the IsoMax loss does not present any drawbacks compared to the SoftMax loss.

The direct replacement of the SoftMax loss by the IsoMax loss significantly improves the baseline OOD detection performance of neural networks. Therefore, rather than the limitations of the models, the low OOD detection performance of deep networks is due to  the SoftMax loss drawbacks, i.e., anisotropy and overconfidence.

In future work, the research community may combine the IsoMax loss with \emph{data-based loss regularization techniques} \cite{NIPS2018_8129, hendrycks2018deep,papadopoulos2019outlier} to improve the performance. \emph{Approaches based on pretrained models} \cite{liang2018enhancing, lee2018simple, Sastry2019DetectingOE} or energy-based fine-tuning/score \cite{DBLP:journals/corr/abs-2010-03759} may be applied on IsoMax loss pretrained networks rather than on SoftMax pretrained models.

Thus, rather than competitors, these approaches are actually \emph{complementary} to IsoMax loss, as they may be \emph{combined} to achieve even higher overall OOD detection performance. IsoMax loss may replace SoftMax loss as a \emph{higher performance baseline for constructing OOD detection solutions}.

\newpage

Another option is to use recent data augmentation techniques \cite{NIPS2019_9540, yun2019cutmix}. We believe that the simplicity of our solution makes it scalable to large images. Hence, we intend to apply this approach to ImageNet \cite{Deng2009ImageNetDatabase}. Finally, since our approach consists of only loss replacement and is based on the general principles of isotropy and maximum entropy, it may be extended to other machine learning methods beyond neural networks.

\ifCLASSOPTIONcaptionsoff
\newpage
\fi

%\vfill\null
%\columnbreak
%\vfill\eject

%\clearpage
\bibliographystyle{ieeetr}
\interlinepenalty=10000
\bibliography{references}

\begin{thebibliography}{10}

\bibitem{DBLP:conf/eccv/WenZL016}
Y.~Wen, K.~Zhang, Z.~Li, and Y.~Qiao, ``A discriminative feature learning
  approach for deep face recognition,'' {\em European Conference on Computer
  Vision}, 2016.

\bibitem{lee2018simple}
K.~Lee, K.~Lee, H.~Lee, and J.~Shin, ``A simple unified framework for detecting
  out-of-distribution samples and adversarial attacks,'' {\em Neural
  Information Processing Systems}, 2018.

\bibitem{Mensink2013DistanceBasedIC}
T.~Mensink, J.~J. Verbeek, F.~Perronnin, and G.~Csurka, ``Distance-based image
  classification: Generalizing to new classes at near-zero cost,'' {\em IEEE
  Transactions on Pattern Analysis and Machine Intelligence}, vol.~35, no.~11,
  pp.~2624--2637, 2013.

\bibitem{Scheirer_2013_TPAMI}
W.~J. Scheirer, A.~Rocha, A.~Sapkota, and T.~E. Boult, ``Towards open set
  recognition,'' {\em {IEEE} Transactions on Pattern Analysis and Machine
  Intelligence}, vol.~35, no.~7, pp.~1757--1772, 2013.

\bibitem{Scheirer_2014_TPAMIb}
W.~J. Scheirer, L.~P. Jain, and T.~E. Boult, ``Probability models for open set
  recognition,'' {\em {IEEE} Transactions on Pattern Analysis and Machine
  Intelligence}, vol.~36, no.~11, pp.~2317--2324, 2014.

\bibitem{7298799}
A.~{Bendale} and T.~{Boult}, ``Towards open world recognition,'' {\em IEEE
  International Conference on Computer Vision and Pattern Recognition}, 2015.

\bibitem{Rudd_2018_TPAMI}
E.~Rudd, L.~P. Jain, W.~J. Scheirer, and T.~Boult, ``The extreme value
  machine,'' {\em {IEEE} Transactions on Pattern Analysis and Machine
  Intelligence}, vol.~40, no.~3, pp.~762--768, 2018.

\bibitem{DeVries2018LearningNetworks}
T.~DeVries and G.~W. Taylor, ``Learning confidence for out-of-distribution
  detection in neural networks,'' {\em CoRR}, vol.~abs/1802.04865, 2018.

\bibitem{Hsu2020GeneralizedOD}
Y.-C. Hsu, Y.~Shen, H.~Jin, and Z.~Kira, ``Generalized {ODIN}: Detecting
  out-of-distribution image without learning from out-of-distribution data,''
  {\em {IEEE} International Conference on Computer Vision and Pattern
  Recognition}, 2020.

\bibitem{Hein2018WhyRN}
M.~Hein, M.~Andriushchenko, and J.~Bitterwolf, ``Why {ReLU} networks yield
  high-confidence predictions far away from the training data and how to
  mitigate the problem,'' {\em {IEEE} International Conference on Computer
  Vision and Pattern Recognition}, 2018.

\bibitem{liang2018enhancing}
S.~Liang, Y.~Li, and R.~Srikant, ``Enhancing the reliability of
  out-of-distribution image detection in neural networks,'' {\em International
  Conference on Learning Representations}, 2018.

\bibitem{NIPS2018_8129}
A.~R. Dhamija, M.~G\"{u}nther, and T.~Boult, ``Reducing network
  agnostophobia,'' {\em Neural Information Processing Systems}, 2018.

\bibitem{hendrycks2018deep}
D.~Hendrycks, M.~Mazeika, and T.~Dietterich, ``Deep anomaly detection with
  outlier exposure,'' {\em International Conference on Learning
  Representations}, 2019.

\bibitem{papadopoulos2019outlier}
A.-A. Papadopoulos, M.~R. Rajati, N.~Shaikh, and J.~Wang, ``Outlier exposure
  with confidence control for out-of-distribution detection,'' {\em CoRR},
  vol.~abs/1906.03509, 2019.

\bibitem{DBLP:journals/corr/abs-2010-03759}
W.~Liu, X.~Wang, J.~D. Owens, and Y.~Li, ``Energy-based out-of-distribution
  detection,'' {\em CoRR}, vol.~abs/2010.03759, 2020.

\bibitem{techapanurak2019hyperparameterfree}
E.~Techapanurak, M.~Suganuma, and T.~Okatani, ``Hyperparameter-free
  out-of-distribution detection using cosine similarity,'' {\em Proceedings of
  the Asian Conference on Computer Vision (ACCV)}, November 2020.

\bibitem{kendall2017uncertainties}
A.~Kendall and Y.~Gal, ``What uncertainties do we need in bayesian deep
  learning for computer vision?,'' {\em Neural Information Processing Systems},
  2017.

\bibitem{Leibig2017LeveragingUI}
C.~Leibig, V.~Allken, M.~S. Ayhan, P.~Berens, and S.~Wahl, ``Leveraging
  uncertainty information from deep neural networks for disease detection,''
  {\em Scientific Reports}, vol.~7, 2017.

\bibitem{subramanya2017confidence}
A.~Subramanya, S.~Srinivas, and R.~V. Babu, ``Confidence estimation in deep
  neural networks via density modelling,'' {\em CoRR}, vol.~abs/1707.07013,
  2017.

\bibitem{malinin2018predictive}
A.~Malinin and M.~Gales, ``Predictive uncertainty estimation via prior
  networks,'' {\em Neural Information Processing Systems}, 2018.

\bibitem{kuleshov2018accurate}
V.~Kuleshov, N.~Fenner, and S.~Ermon, ``Accurate uncertainties for deep
  learning using calibrated regression,'' {\em International Conference on
  Machine Learning}, 2018.

\bibitem{liu2016large}
W.~Liu, Y.~Wen, Z.~Yu, and M.~Yang, ``Large-margin softmax loss for
  convolutional neural networks.,'' {\em International Conference on Machine
  Learning}, 2016.

\bibitem{Guo2017OnCO}
C.~Guo, G.~Pleiss, Y.~Sun, and K.~Q. Weinberger, ``On calibration of modern
  neural networks,'' {\em International Conference on Machine Learning}, 2017.

\bibitem{PhysRev.106.620}
E.~T. Jaynes, ``Information theory and statistical mechanics,'' {\em Physical
  Review}, vol.~106, pp.~620--630, 1957.

\bibitem{PhysRev.108.171}
E.~T. Jaynes, ``Information theory and statistical mechanics. {II},'' {\em
  Physical Review}, vol.~108, pp.~171--190, 1957.

\bibitem{10.5555/1146355}
T.~M. Cover and J.~A. Thomas, ``Elements of information theory,'' {\em Wiley
  Series in Telecommunications and Signal Processing}, 2006.

\bibitem{shafaei2018biased}
A.~Shafaei, M.~Schmidt, and J.~J. Little, ``A less biased evaluation of
  out-of-distribution sample detectors,'' {\em British Machine Vision
  Conference}, 2019.

\bibitem{glorot2010understanding}
X.~Glorot and Y.~Bengio, ``Understanding the difficulty of training deep
  feedforward neural networks,'' {\em International Conference on Artificial
  Intelligence and Statistics}, 2010.

\bibitem{He2016DelvingClassification}
K.~He, X.~Zhang, S.~Ren, and J.~Sun, ``{Delving deep into rectifiers:
  Surpassing human-level performance on imagenet classification},'' {\em
  International Conference on Computer Vision}, 2016.

\bibitem{DBLP:conf/nips/RenLFSPDDL19}
J.~Ren, P.~J. Liu, E.~Fertig, J.~Snoek, R.~Poplin, M.~A. DePristo, J.~V.
  Dillon, and B.~Lakshminarayanan, ``Likelihood ratios for out-of-distribution
  detection,'' {\em Neural Information Processing Systems}, 2019.

\bibitem{DBLP:journals/spl/WangCLL18}
F.~Wang, J.~Cheng, W.~Liu, and H.~Liu, ``Additive margin softmax for face
  verification,'' {\em {IEEE} Signal Processing Letters}, vol.~25, no.~7,
  pp.~926--930, 2018.

\bibitem{DBLP:conf/cvpr/DengGXZ19}
J.~Deng, J.~Guo, N.~Xue, and S.~Zafeiriou, ``{ArcFace: A}dditive angular margin
  loss for deep face recognition,'' {\em IEEE International Conference on
  Computer Vision and Pattern Recognition}, 2019.

\bibitem{Deng2009ImageNetDatabase}
J.~Deng, W.~Dong, R.~Socher, L.~Li, K.~Li, and F.~Li, ``{ImageNet: A
  large-scale hierarchical image database},'' {\em {IEEE} International
  Conference on Computer Vision and Pattern Recognition}, 2009.

\bibitem{Yu2015LSUNLoop}
F.~Yu, Y.~Zhang, S.~Song, A.~Seff, and J.~Xiao, ``{LSUN:} construction of a
  large-scale image dataset using deep learning with humans in the loop,'' {\em
  CoRR}, vol.~abs/1506.03365, 2015.

\bibitem{hendrycks2017baseline}
D.~Hendrycks and K.~Gimpel, ``A baseline for detecting misclassified and
  out-of-distribution examples in neural networks,'' {\em International
  Conference on Learning Representations}, 2017.

\bibitem{Huang2017DenselyNetworks}
G.~Huang, Z.~Liu, L.~v.~d. Maaten, and K.~Q. Weinberger, ``Densely connected
  convolutional networks,'' {\em {IEEE} International Conference on Computer
  Vision and Pattern Recognition}, 2017.

\bibitem{He_2016}
K.~He, X.~Zhang, S.~Ren, and J.~Sun, ``Identity mappings in deep residual
  networks,'' {\em European Conference on Computer Vision}, 2016.

\bibitem{Krizhevsky2009LearningImages}
A.~Krizhevsky, ``Learning multiple layers of features from tiny images,'' {\em
  Science Department, University of Toronto}, 2009.

\bibitem{Netzer2011ReadingLearning}
Y.~Netzer and T.~Wang, ``{Reading digits in natural images with unsupervised
  feature learning},'' {\em Neural Information Processing Systems}, 2011.

\bibitem{DBLP:conf/iclr/TsiprasSETM19}
D.~Tsipras, S.~Santurkar, L.~Engstrom, A.~Turner, and A.~Madry, ``Robustness
  may be at odds with accuracy,'' {\em 7th International Conference on Learning
  Representations, {ICLR} 2019, New Orleans, LA, USA, May 6-9, 2019}, 2019.

\bibitem{carlini2019evaluating}
N.~Carlini, A.~Athalye, N.~Papernot, W.~Brendel, J.~Rauber, D.~Tsipras, I.~J.
  Goodfellow, A.~Madry, and A.~Kurakin, ``On evaluating adversarial
  robustness,'' {\em CoRR}, vol.~abs/1902.06705, 2019.

\bibitem{Schwartz2019GreenA}
R.~Schwartz, J.~Dodge, N.~A. Smith, and O.~Etzioni, ``Green {AI},'' {\em CoRR},
  vol.~abs/1907.10597, 2019.

\bibitem{DBLP:conf/nips/ShafahiNG0DSDTG19}
A.~Shafahi, M.~Najibi, A.~Ghiasi, Z.~Xu, J.~P. Dickerson, C.~Studer, L.~S.
  Davis, G.~Taylor, and T.~Goldstein, ``Adversarial training for free!,'' {\em
  Neural Information Processing Systems}, 2019.

\bibitem{Sastry2019DetectingOE}
C.~S. Sastry and S.~Oore, ``Detecting out-of-distribution examples with gram
  matrices,'' {\em International Conference on Machine Learning}, vol.~119,
  pp.~8491--8501, 2020.

\bibitem{NIPS2019_9540}
S.~Thulasidasan, G.~Chennupati, J.~A. Bilmes, T.~Bhattacharya, and S.~Michalak,
  ``On mixup training: Improved calibration and predictive uncertainty for deep
  neural networks,'' {\em Neural Information Processing Systems}, 2019.

\bibitem{yun2019cutmix}
S.~Yun, D.~Han, S.~J. Oh, S.~Chun, J.~Choe, and Y.~Yoo, ``Cutmix:
  Regularization strategy to train strong classifiers with localizable
  features,'' {\em International Conference on Computer Vision}, 2019.

\end{thebibliography}

\end{document}